\documentclass[letterpaper, 10 pt, conference]{ieeeconf}  

\IEEEoverridecommandlockouts                              

\usepackage{graphics} 
\usepackage{epsfig} 
\usepackage{times} 
\usepackage[T1]{fontenc}
\usepackage[ruled,vlined,linesnumbered]{algorithm2e}

\usepackage{amsmath, graphicx}
\usepackage{balance}
\usepackage{amssymb,amsbsy}
\usepackage{breqn,bbm}
\usepackage[dvipsnames]{xcolor}
\usepackage{multirow}
\usepackage[symbol,bottom]{footmisc}
\usepackage{soul}
\usepackage{url}
\usepackage{booktabs} 
\usepackage{booktabs}
\usepackage{tikz}
\usepackage[export]{adjustbox}
\usepackage{pifont}
\usepackage{subcaption}
\usepackage{float}
\usepackage{tabularx}
\usepackage{amssymb}
\usepackage{booktabs}
\usepackage{pifont}

\SetCommentSty{mycommfont}

\title{\LARGE \bf
Efficient Surgical Robotic Instrument Pose Reconstruction in Real World Conditions Using Unified Feature Detection
}

\author{Zekai Liang$^{1}$, Kazuya Miyata$^{1}$, Xiao Liang$^{1}$, Florian Richter$^{1}$, Michael C. Yip$^{1}$, \IEEEmembership{Senior Member, IEEE} 
\thanks{$^{1}$Department of Electrical and Computer Engineering, University of California San Diego, La Jolla, CA 92093 USA.\ {\tt\small \{z9liang, kamiyata, x5liang, frichter, yip\}@ucsd.edu}}
}

\begin{document}

\maketitle
\thispagestyle{empty}
\pagestyle{empty}

\begin{abstract}
Accurate camera-to-robot calibration is essential for any vision-based robotic control system and especially critical in minimally invasive surgical robots, where instruments conduct precise micro-manipulations.
However, MIS robots have long kinematic chains and partial visibility of their degrees of freedom in the camera, which introduces challenges for conventional camera-to-robot calibration methods that assume stiff robots with good visibility. Previous works have investigated both keypoint-based and rendering-based approaches to address this challenge in real-world conditions;
however, they often struggle with consistent feature detection or have long inference times, neither of which are ideal for online robot control.
In this work, we propose a novel framework that unifies the detection of geometric primitives (keypoints and shaft edges) through a shared encoding, enabling efficient pose estimation via projection geometry. This architecture detects both keypoints and edges in a single inference and is trained on large-scale synthetic data with projective labeling.
This method is evaluated across both feature detection and pose estimation, with qualitative and quantitative results demonstrating fast performance and state-of-the-art accuracy in challenging surgical environments. \textit{The code will be released upon paper acceptance.}

\end{abstract}

\section{Introduction}

In recent years, autonomous robotic-assisted Minimal-Invasive-Surgery (MIS) has drawn increasing attention for its efficiency and safety, and  reducing surgeons' workload and fatigue from long-time operations.
Engineering solutions to aid during MIS such as augmented reality guidance \cite{seetohul2023augmented} or task automation \cite{shkurti2025systematic}, require accurate surgical instrument localization to provide precise and safe assistance.

Modern vision-based robot pose estimation works have been proposed in recent years, which can be generally categorized in two paradigms: keypoint-based \cite{lambrecht2019towards, lee2020camera, lu2025ctrnet} and rendering-based \cite{labbe2021single, lu2023image} methods. 
Surgical robots like the da Vinci system from Intuitive, however, utilize long thin instruments with cable-driven transmissions to enable smooth motions at distal locations. Such mechanisms introduce a combination of long-chain kinematic errors, compliant bending, and cable nonlinearities that cumulatively result in significant end-effector pose errors that cannot be measured in the robot joints. 
Additionally, in laparoscopic surgery, the limited camera view restricts access to the full kinematic chain, leading to partial visibility and degraded video quality. These challenges set surgical robots apart from traditional pose estimation and make accurate tool tracking especially difficult.

Previous studies have explored keypoint detection~\cite{lu2021super}, optimal keypoint placement~\cite{lu2022pose}, and differentiable rendering-based matching~\cite{liang2025differentiable} to address this challenge. Keypoint-based methods typically rely on a Perspective-n-Point (PnP) solver to estimate the pose with the detected points and kinematic information. Nevertheless, in surgical robotics, even the most recent keypoint methods are often unreliable due to low video quality, frequent occlusions, and the small scale of the instruments. At the same time, rendering-based approaches achieve more robustness and consistency by direct contour matching, but they are still constrained by long processing times as they require an online  iterative alignment process; they also typically require clear contours in view, and are susceptible to convergence to incorrect local minima during optimization.



\begin{figure}[t]
    \centering
    \includegraphics[width=0.95\linewidth,clip=true,trim={0mm 5mm 5mm 0mm}]{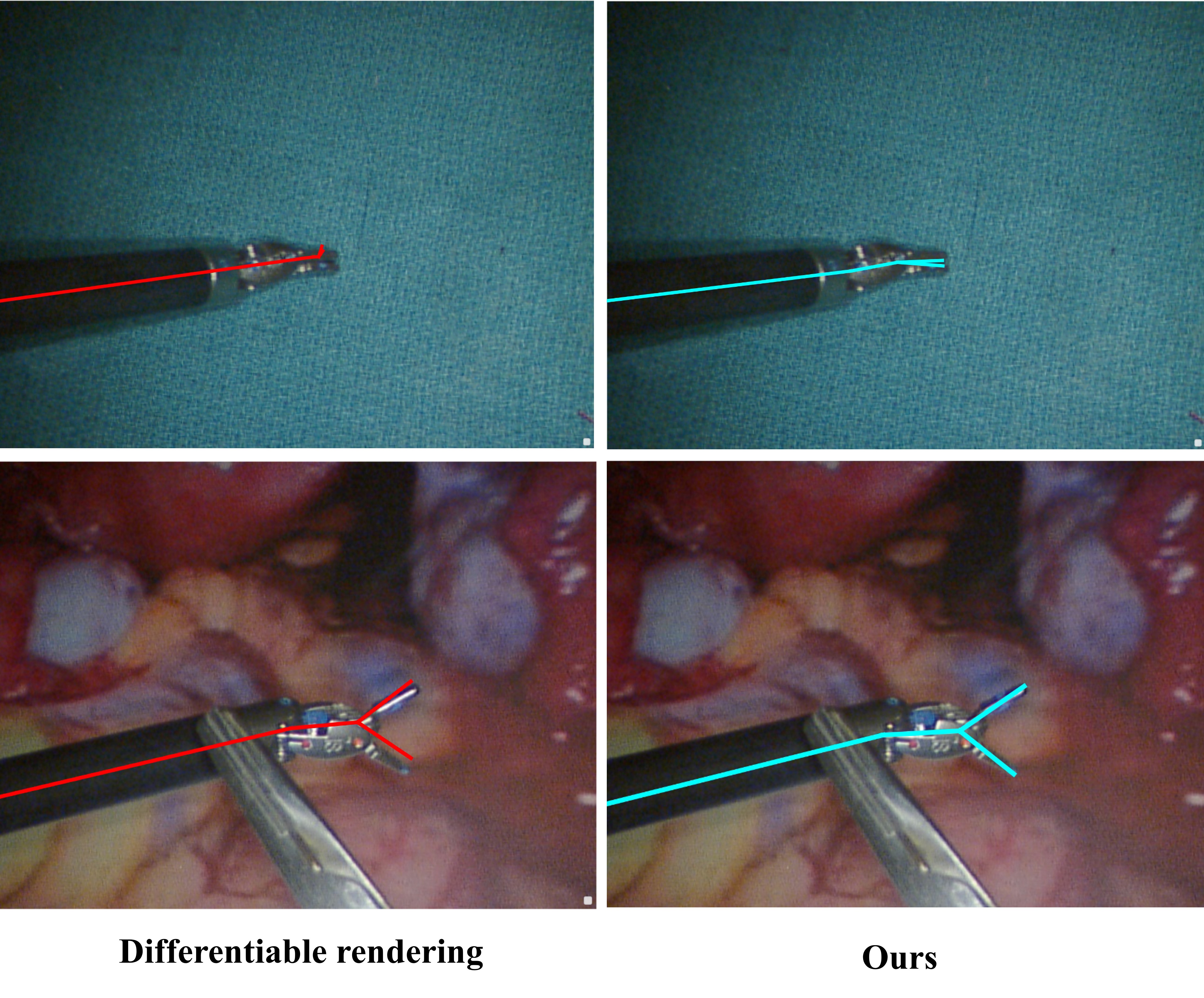}
    \caption{Pose reconstruction comparison between our framework and differentiable rendering based method. The skeleton overlay is obtained by estimated pose and forward kinematics.}
    \label{vs diff}
    \vspace{-0.14in}
\end{figure}

\begin{figure*}[ht]
    \centerline{\includegraphics[width=0.87\linewidth,clip=true,trim={0mm 0mm 0mm 0mm}]{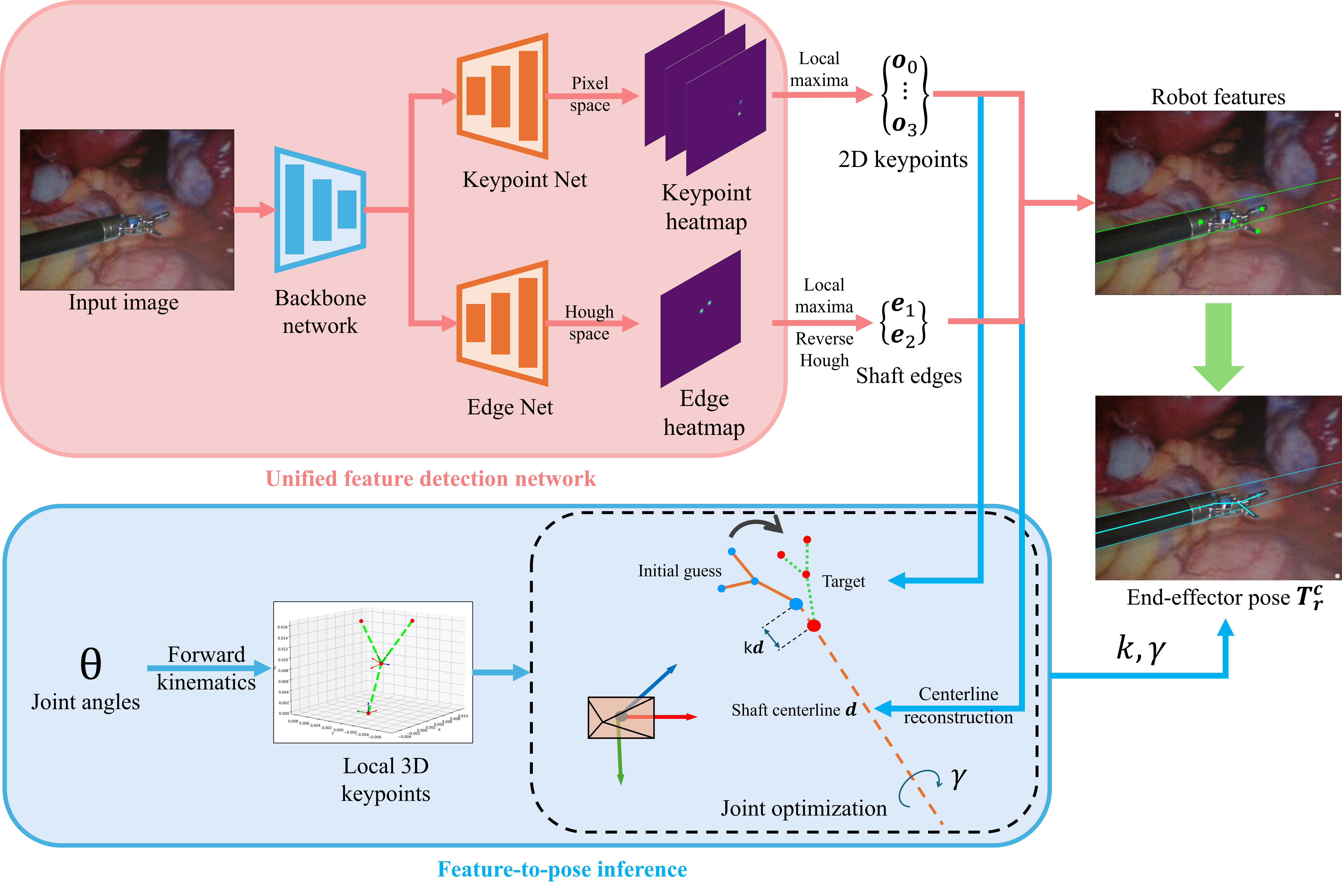}}
    \caption{The overview of the proposed framework. Keypoint Net and Edge Net are jointly trained on large-scale synthetic data using heatmap regression with a shared encoder. During inference, the detected keypoints and shaft edges are passed to a geometric pose solver, which leverages the robot’s projective constraints to efficiently estimate the full 6D pose.}
    \label{pipeline}
    \vspace{-0.14in}
    
\end{figure*}

To address these limitations, we propose a unified framework for fast surgical instrument pose estimation that integrates the strengths of both keypoint-based and rendering-based paradigms while avoiding their respective drawbacks through a direct geometric formulation. Specifically, the proposed method treats shaft edges as a learnable geometric primitive trained jointly with keypoint detection on a large-scale, realistically randomized synthetic dataset to mitigate the sim-to-real gap. At inference time, the detected keypoints and shaft edges are combined with known kinematic priors of the surgical robot arm, enabling efficient feature-to-pose estimation without reliance on iterative postprocessing. This framework is evaluated both qualitatively and quantitatively, demonstrating significant improvements in feature detection and pose reconstruction compared with prior approaches.




\section{Related work}

Accurate robot pose estimation from visual input has long been a key requirement for vision-based control.
Keypoint-based approaches~\cite{lambrecht2019towards,lee2020camera,lu2022pose,lu2021super} detect robot landmarks and recover pose with a PnP solver, whereas rendering-based methods~\cite{labbe2021single,lu2023image} align projected models to image observations. Hybrid frameworks, such as CtRNet~\cite{lu2023markerless,lu2025ctrnet}, combine both paradigms to enable self-supervised training on unlabeled real data. Although these approaches have achieved notable progress, their applicability to surgical robotics remains limited due to the complexity of surgical scenes and the unconventional design of surgical manipulators.

In surgical settings, 
long serial-chain transmissions, flexible shafts, and cable-driven actuation introduce significant unmeasured nonlinearities into the true kinematics of the robots. Previous methods attempted to compensate by modeling cable stretch and friction~\cite{miyasaka2015measurement}, learning end-effector offsets~\cite{mahler2014learning,seita2018fast,pastor2013learning}, or calibrating the remote center of motion (RCM)~\cite{zhong2020hand,zhao2015efficient, lu2022unified, li2024real}. Deep learning has also been applied to markerless pose estimation~\cite{lu2022pose,reiter2012feature,fan2024reinforcement}. However, the endoscopic cameras provide only a narrow field of view, limited resolution, and suboptimal lighting, making feature detection, particularly of keypoints, highly error-prone. Richter et al. \cite{richter2021robotic} proposed a lumped-error formulation that combined spatio-temporal consistency in a particle filter approach to track the robot pose, incorporating the instrument shaft as a robust geometric primitive under complex surgical conditions. d'Ambrosia et al. \cite{d2024robust} further improved this observation model with neural networks to enhance edge detection. Despite recognizing the importance of shaft edges, existing pipelines still extract shaft edges at the contour level, either by selecting the longest lines from Canny–Hough transforms or by performing image-pair matching, which represents a non-learnable design that often performs poorly in cluttered and noisy surgical scenes.


More recently, \cite{liang2025differentiable} proposes a differentiable rendering framework that enforces geometric constraints to achieve robust frame-level pose estimation, eliminating the need for manual correspondences and painted markers. While this approach substantially improves robustness, rendering-based methods still suffer from long optimization times, rely on segmentation methods that can produce incorrect masks, require fully clear contours, and have many incorrect local minima solutions.


\section{Methodology}
The complete inference pipeline of this framework is illustrated in Fig. \ref{pipeline}. This presents the first solution that unifies the feature detection of surgical robots into a single neural network with heatmap regression, elevating the shaft edges into a crucial but learnable feature as keypoints. To enable large-scale training without the excessive burden of manual labeling, the state-of-the-art simulation engine with photorealistic rendering and feature projection is leveraged to generate synthetic data efficiently. Furthermore, a geometric pose solver that utilizes projective constraints is introduced, achieving fast and robust 6D pose estimation. 



\vspace{-0.1in}
\subsection{Training data generation}
\label{sec:data}
Synthetic data enables large-scale training without the time-consuming manual annotation, while providing consistent and precise ground-truth labels.
The synthetic data generation pipeline is set up in Isaac Sim from NVIDIA Omniverse, supporting high-quality rendering that closely matches real-world images.
In real surgical robot operation scenes, operators have very limited visibility of the full kinematic chain.  Multiple domain randomization steps are applied to randomize the instrument pose and shaft configuration in the image, lighting conditions, and image background: (1) initialize the camera to base transformation with a reference transformation from a real-world setup. (2) a random rotation is applied to the camera about its depth axis by an angle uniformly sampled from $[-\pi, \pi]$. (3) a random visible end-effector pose is sampled based on the camera view. Each end-effector pose sample is constrained to be within $[z_0, z_1]$ mm in the direction of the camera's depth axis. (4) the kinematic feasibility of the sampled end-effector pose is checked; if not feasible, steps 3 and 4 are repeated. (5) randomly sample the grippers joint angle. (6) lighting parameters are randomized and scene backgrounds are sampled from IsaacSim's replicator assets.

To efficiently generate large-scale training data with ground truth annotation, the cylinder projection from \cite{richter2021robotic} is utilized to generate the ground truth shaft edges in the equation form, $Au + Bv + C = 0$, where $u,v$ are pixel coordinates and $A,B,C$ are the projected edge parameters. The edge parameters are computed as 
    \begin{equation}
        \small
        \begin{aligned}
        A_{1,2} &=
        \frac{r \left( x_0^c - a^c (\mathbf{p}_0^c)^\top \mathbf{d}^c \right)}
        {\sqrt{(\mathbf{p}_0^c)^\top \mathbf{p}_0^c - (\mathbf{p}_0^c)^\top \mathbf{d}^c - r^2}}
        \pm (c^c y_0^c - b^c z_0^c) 
        \\
        B_{1,2} &=
        \frac{r \left( y_0^c - b^c (\mathbf{p}_0^c)^\top \mathbf{d}^c \right)}
        {\sqrt{(\mathbf{p}_0^c)^\top \mathbf{p}_0^c - (\mathbf{p}_0^c)^\top \mathbf{d}^c - r^2}}
        \pm (a^c z_0^c - c^c x_0^c) 
        \\
        C_{1,2} &=
        \frac{r \left( z_0^c - c^c (\mathbf{p}_0^c)^\top \mathbf{d}^c \right)}
        {\sqrt{(\mathbf{p}_0^c)^\top \mathbf{p}_0^c - (\mathbf{p}_0^c)^\top \mathbf{d}^c - r^2}}
        \pm (b^c x_0^c - a^c y_0^c)
        \end{aligned}
        \label{ABC}
    \end{equation}
where $\textbf{p}_0^c = [x_0^c, y_0^c, y_0^c]$ is a point on the center line of the insertion shaft (i.e. cylinder) being projected, $\textbf{d}^c = [a^c, b^c, c^c]$ is the center line direction, and $r$ is the radius of the insertion shaft.
The insertion shaft of a surgical laparoscopic tool will practically always be present in the camera view when the instrument is visible and provides a strong signal for localization.

\begin{figure}[t]
    \centerline{\includegraphics[width=0.90\linewidth,clip=true,trim={0mm 0mm 0mm 0mm}]
    {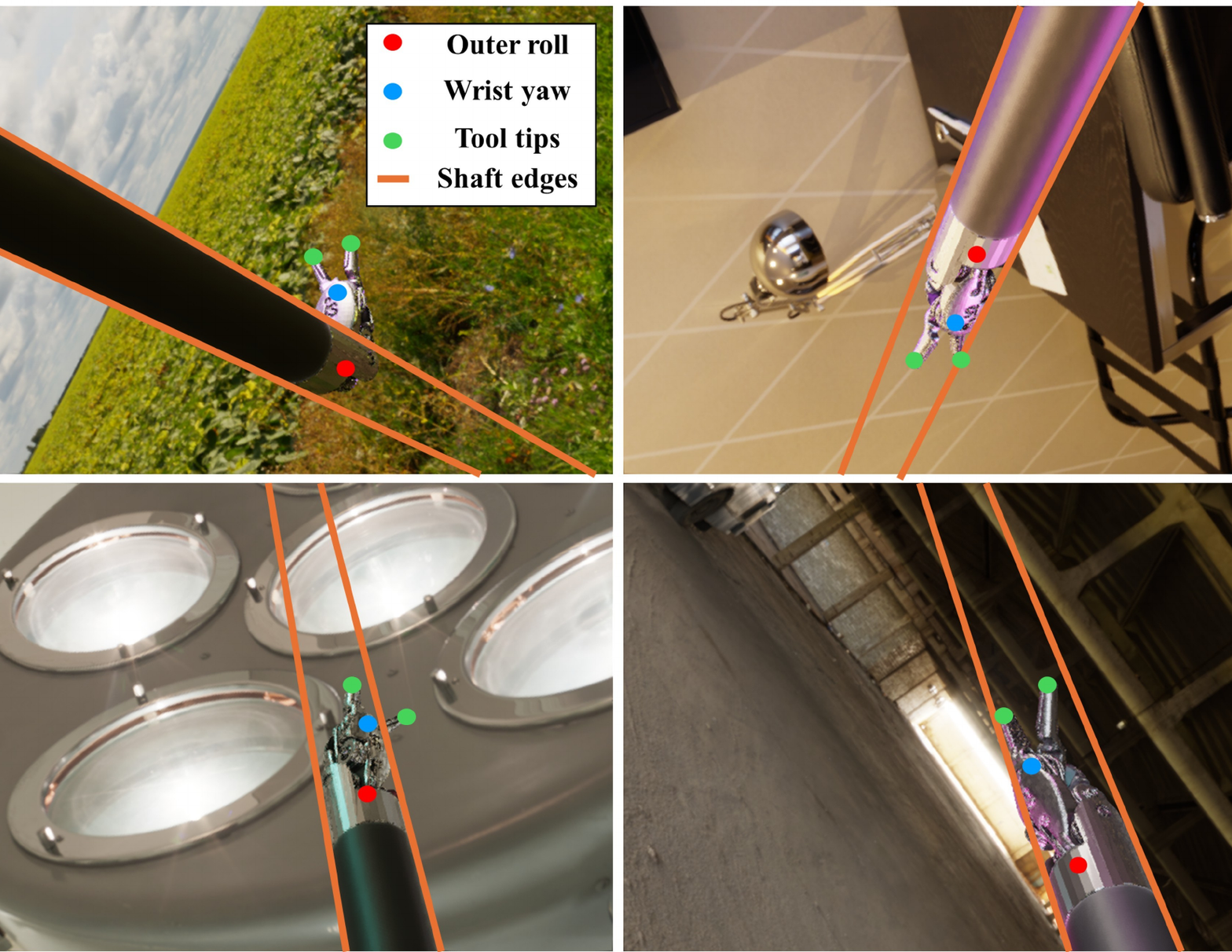}}
    \caption{Synthetic training data generated with ground truth shaft edges and keypoint annotations (Outer Roll, Wrist Yaw and Tool Tips).}
    \label{training data}
    \vspace{-0.15in}
\end{figure}

Keypoints, on the other hand, can be noisy to detect due to occlusion, debris, poor lighting, smoke, and other suboptimal visual conditions, but provide wrist features that are key to reconstructing the end-effector orientation. A total of 4 target keypoints are placed at the two Tool Tips and the last two robot joint frames, Outer Roll and Wrist Yaw, to reduce the complexity of noisy keypoint detection. The ground truth 2D keypoints are obtained using the pinhole model projection. Our sample data and corresponding annotation are shown in Fig. \ref{training data}. 


\subsection{Unified feature detection network}
\label{sec:network}

\subsubsection{Model overview} 
As shown in Fig. \ref{pipeline}, this framework adopts a unified architecture that jointly predicts line features in Hough space and keypoint locations in the pixel space from a shared backbone network. 
In surgical scenes, the kinematic chain of the surgical tools is partially visible and the portion of the tool within the camera view frequently becomes occluded, which makes predicting the endpoints of the shaft edges and other line representations in pixel space challenging and suboptimal.
Inspired by Deep Hough Transform (DHT) \cite{zhao2021deep}, which uses polar form $(\rho, \theta)$ as a global representation of lines, the shaft edges are transformed from pixel space to Hough space: 


\vspace{-0.1in}
\begin{equation}
    \footnotesize
    \begin{aligned}
    \rho   &= \frac{1}{\sqrt{(A/(-C))^2 + (B/(-C))^2}}, \quad
    \theta = \operatorname{atan2}(B/(-C), A/(-C))
    \end{aligned}
\end{equation}

\begin{equation}
    \footnotesize
    \theta_c = 
    \begin{cases}
    \theta, & \theta \ge 0 \\
    \theta + \pi, & \theta < 0
    \end{cases}, \quad
    \rho' = 
    \begin{cases}
    \rho, & \theta \ge 0 \\
    -\rho, & \theta < 0
    \end{cases}
\end{equation}

\begin{equation}
    \rho_c = \rho' - \frac{W}{2} \cos\theta_c - \frac{H}{2} \sin\theta_c
\end{equation}

\begin{equation}
    \rho_c = u\cos\theta_c + v\sin\theta_C
\end{equation}
where standard line parameters $(Au + Bv + C = 0)$ are the input. By shifting the transform to the image center, each line can be represented with a unique pair of  $\theta_c \in [0, \pi]$ and $\rho_c \in \left[-\tfrac{\sqrt{W^2+H^2}}{2}, \tfrac{\sqrt{W^2+H^2}}{2}\right]$, where $W$ and $H$ are the input image dimensions.

The foundation model DINOv2-L \cite{oquab2023dinov2} is used as the backbone network due to its strong generalization capability across domains. 
For an input RGB image of size $224 \times 224$, the ViT architecture with a patch size of $14 \times 14$ divides the image into a $16 \times 16$ grid of patches, resulting in $N=256$ patch tokens. The backbone outputs patch-level embeddings 
$\mathbf{F} \in \mathbb{R}^{B \times N \times D}$, 
where $B$ is the batch size and $D$ is the hidden dimension of the backbone. The patch tokens are reshaped into a spatial feature map $\mathbf{F}_{map} \in \mathbb{R}^{B \times D \times 16 \times 16}$. 

For edge detection, a lightweight CNN-based Edge Net that progressively increases spatial resolution is adopted, projecting the $16\times16$ backbone features onto a dense $180\times180$ Hough space grid.  
This head consists of a total of 4 up-blocks, each composed of one bilinear up-sampling by a factor of two, followed by two convolution + ReLU layers. Four such stages expand the feature map from $16\times16$ to $256\times256$. Finally, a $1\times1$ convolution followed by bilinear resizing produces the logits on the target $180\times180$ grid. Since on the image plane the two shaft edges are symmetric and identical, they are included in the same channel of the output.

The Keypoint Net shares the same upsampling strategy, refining the $16\times16$ backbone feature map progressively to a high-resolution heatmap of size $256{\times}256$. Then, a $1{\times}1$ convolution is applied to obtain $C_{\text{kpt}}$ channels, followed by bilinear resizing to the exact image resolution ($224{\times}224$). Each output channel corresponds to a certain target keypoint, while the last two keypoints on the tool tips share the same output channel due to symmetric ambiguity.

\subsubsection{Network training} 
Following standard heatmap regression techniques which are extensively used in the pose estimation tasks \cite{zhang2020distribution, xu2023vitpose++, khirodkar2024sapiens}, line annotations are discretized on a $180 \times 180$ grid, where each bin corresponding to a line is smoothed with a Gaussian kernel and normalized to $[0,1]$.
Keypoints are similarly projected into the $224 \times 224$ image plane, placed as impulses in separate channels, and Gaussian-blurred to form smooth supervision signals.

The network is jointly trained on synthetic data with both keypoint and line supervision. To handle the highly imbalanced distribution of foreground peaks and background pixels in the heatmaps, the Adaptive Wing loss~\cite{wang2019adaptive} is applied for both the keypoint and line heads. This loss adaptively sharpens the penalty around peak regions while relaxing it in smooth background areas, encouraging the model to produce accurate and well-localized responses. The total training objective is formulated as a weighted sum of the edge and keypoint losses:

\vspace{-0.1in}
\begin{equation}
    \small
    \mathcal{L}_{\text{AWing}}(y, y') = 
    \begin{cases}
    \omega \, \ln \!\big( 1 + |y-y'|^{\alpha - y'} \big), & |y-y'| < \phi, \\
    \eta \, |y-y'| - \nu, & |y-y'| \geq \phi,
    \end{cases}
\end{equation}

\begin{equation}
\mathcal{L} = \lambda_{\text{line}} \, \mathcal{L}_{\text{AWing}}^{\text{line}}
            + \lambda_{\text{kpt}} \, \mathcal{L}_{\text{AWing}}^{\text{kpt}},
\end{equation}
where $y$ and $y'$ are the per-bin heatmap values for the ground truth and prediction results. $\lambda_{\text{line}}$ and $\lambda_{\text{kpt}}$ are the scaling factors of two branches, and $\alpha$, $\omega$, $\phi$, $\eta$, and $\nu$ are hyperparameters of the loss function.

\begin{algorithm}[t]
    \footnotesize
    \caption{Pixel Level Edge Refinement}
    \label{alg: refinement}
    \KwIn{Input image $\mathbb{I}$, initial line parameters $(A, B, C)$, distance threshold $d$}
    \KwOut{Refined line parameters $(A', B', C')$}

    
    $\mathbb{E} \leftarrow \text{LineSegmentDetector}(\mathbb{I})$
    
    \ForEach{pixel $(x, y)$ where $\mathbb{E}(x, y)$ is an edge}{
        \If{$\left|\frac{Ax + By + C}{\sqrt{A^2 + B^2}}\right| < d$}{
            Inlier set $\mathcal{P} \leftarrow (x, y)$ 
        }
    }
    
    \If{$|\mathcal{P}| < 10$}{
        \Return $(A, B, C)$
    }
    Fit line $y = mx + b$ to $\mathcal{P}$ using RANSAC
    
    $(A', B', C') \leftarrow (-m, 1, -b)$ \\
    
    \Return $(A', B', C')$
   
\end{algorithm}

\begin{figure}[t]
    \centerline{\includegraphics[width=1\linewidth,clip=true,trim={10mm 10mm 10mm 10mm}]{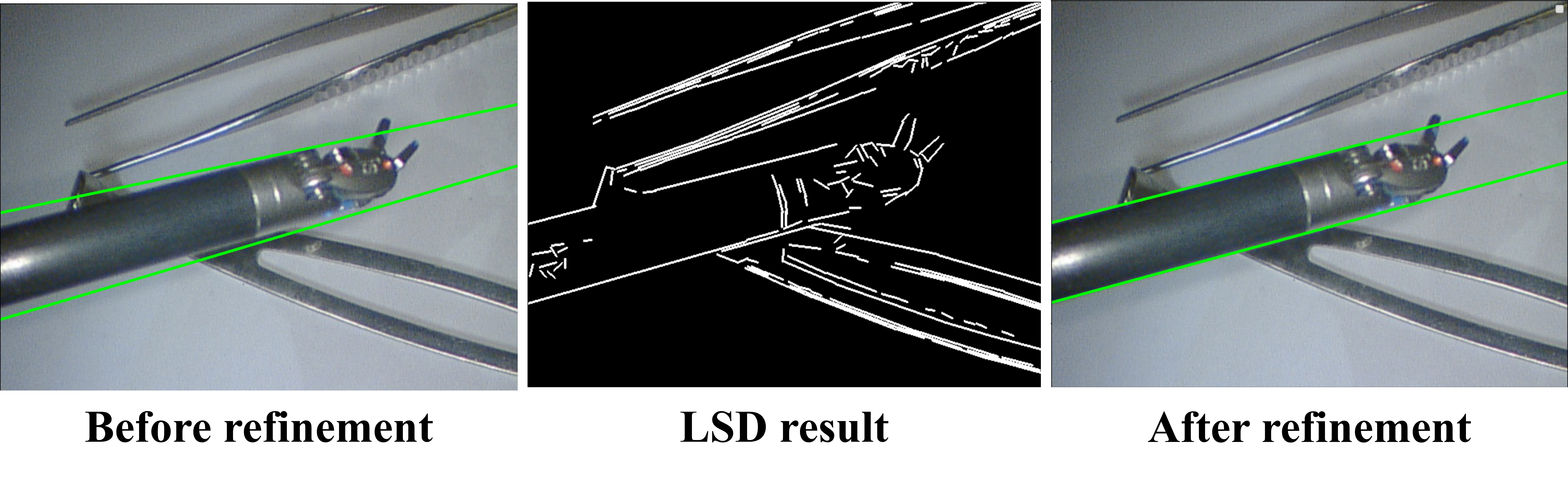}}
    \caption{We apply a pixel-level edge refinement to the output of Edge Net using  Line Segment Detector to achieve a more accurate shaft estimation.}
    \label{refinement}
    \vspace{-0.14in}
\end{figure}

\subsection{Feature-to-pose inference}
\label{sec:solver}

Based on the network output, we propose a fast feature-to-pose pipeline that achieves fast and robust performance in real-world conditions. 
As outlined in Algorithm \ref{alg:inference}, the pose solver generally consists of 2 steps: reconstructing the shaft orientation and solving the shaft roll with robust feature matching.
For the sake of simplicity in our explanation and equations, we will be considering the end-effector at the end of the insertion shaft before the gripper starts.
This corresponds with joint 4 on the dVRK \cite{kazanzides2014open} and the Outer Roll keypoint shown in Fig. \ref{training data}.
The proposed approach still considers the entire gripper and forward kinematics can be applied to transform the resulting pose to the grippers coordinate frame.

\subsubsection{Feature extraction} For inference, heatmaps are decoded by extracting local maxima similar to prior pose estimation approaches \cite{zhang2020distribution, xu2023vitpose++, khirodkar2024sapiens}. Keypoint heatmaps yield peak pixel coordinates per channel $\textbf{u}_i$, while line heatmaps are decoded into parametric $(\theta_i, \rho_i)$ representations by selecting top-scoring peaks. For computation in later steps, the inverse Hough transform is further applied, converting line parameters to $\textbf{e}_i = (A_i, B_i, C_i)$, where $A_iu + B_iv + C_i = 0$.

Due to the sensitivity of polar line representations, slight disturbances in the parameters may result in substantial shifts in the line’s position. To minimize the noise introduced by heatmap prediction, a light-weight refinement module is applied to better align the estimated lines to pixel-level image edges. As shown in Algorithm \ref{alg: refinement}, Line Segment Detector \cite{von2008lsd} is utilized to generate sparse edge maps $\mathbb{E}$, while the positive edge pixels within a distance threshold $d$ to the original lines are included in the inlier set $\mathcal{P}$. Finally, refined line parameters $(A', B', C')$ are obtained using RANSAC fitting.

\subsubsection{Shaft centerline reconstruction} Following \cite{doignon2007degenerate} and later works, the cylinder’s 3D position and orientation in space $\textbf{a},\textbf{d}  \in \mathbb{R}^3$ can be recovered given the actual radius $r$ and two edges of a projected cylinder 
$\textbf{e}_{1}, \textbf{e}_{2} \in \mathbb{R}^3$:

\begin{equation}
    \hat{\textbf{a}} = \frac{\textbf{v}^{+}}{\|\textbf{v}^{+}\|}, 
    \qquad 
    \textbf{d} = \frac{\textbf{v}^{-}}{\|\textbf{v}^{-}\|} \times \hat{\textbf{a}}
\end{equation}
where
\begin{equation}
    \small
    \textbf{v}^{+} = \tfrac{1}{2}\left(\tfrac{\textbf{e}_{1}}{\|\textbf{e}_{1}\|} + \tfrac{\textbf{e}_{2}}{\|\textbf{e}_{2}\|}\right), \quad
    \textbf{v}^{-} = \tfrac{1}{2}\left(\tfrac{\textbf{e}_{1}}{\|\textbf{e}_{1}\|} - \tfrac{\textbf{e}_{2}}{\|\textbf{e}_{2}\|}\right).
\end{equation}

\noindent The cylinder position here stands for the closest point from the centerline to the camera in space, $\textbf{a} =\|a\| \hat{\textbf{a}}$, and here, $\hat{(\cdot)}$ denotes a unit vector. The magnitude $\|a\|$ of the position vector is obtained from the inner product of edges:
\begin{equation}
    \|a\| = r \sqrt{\tfrac{2}{\,1+ \textbf{e}_{1}^\top \textbf{e}_{2} \,/\, (\|\textbf{e}_{1}\|\|\textbf{e}_{2}\|)}}
\end{equation}

\begin{algorithm}[t]
    \footnotesize
    \caption{Feature-to-Pose Inference}
    \label{alg:inference}
    \KwIn{Image $\mathbb{I}$, joint angles $\mathbf{q}$}
    \KwOut{Camera-to-End-Effector transform $\mathbf{T}_{\mathrm{cam}\rightarrow\mathrm{ee}}$}

    \CommentSty{// Network Inference} \\
    $\mathbb{H}_{\text{edge}}, \mathbb{H}_{\text{kpt}} \gets \text{Model}(\mathbb{I})$ \\

    \CommentSty{// Extract edges and keypoints} \\
    $\textbf{e}_i \gets \text{findLocalMaxima}(\mathbb{H}_{\text{edge}}),\ i\in\{1,2\}$ \\
    $\textbf{e}_i \gets \text{pixelLevelRefinement}(\textbf{e}_i, \mathbb{I}) ,\ i\in\{1,2\}$ \\
    $\mathbf{u}_i \gets \text{findLocalMaxima}(\mathbb{H}_{\text{kpt}}),\ i\in\{0,1,2,3\}$ \\

    \CommentSty{// Cylinder inversion} \\
    $(\mathbf{a},\mathbf{d}) \gets \text{InvertCylinder}(\textbf{e}_1,\;\textbf{e}_2)$ \\

    \CommentSty{// Recover initial position} \\
    $\mathbf{p}_0 \gets \text{RecoverPoint3D}(\mathbf{K},\mathbf{a},\mathbf{d},\mathbf{u}_0)$ \\

    \CommentSty{// Forward kinematics } \\
    $\{\mathbf{x}_j\}_{j=1}^{3} \gets \text{FK}(\mathbf{q})$ \\

    \CommentSty{// Recover initial rotation} \\
    $\mathbf{R}_{\text{align}} \gets \text{AlignRotation}(\mathbf{e}_z,\hat{\mathbf{d}})$ \\ 
    \CommentSty{// Pose parameterization} \\
    $\mathbf{R}_{ee}(\gamma) \gets \mathbf{R}_{\text{align}}\mathbf{R}_z(\gamma)$ \\
    $\mathbf{t}_{ee}(k) \gets \mathbf{p}_0 + k\,\hat{\mathbf{d}}$ \\

    \CommentSty{// Reprojection residual} \\
    $\hat{\mathbf{u}}_j(\gamma,k) \gets \pi(\mathbf{K}[\,\mathbf{R}_{ee}(\gamma)\mathbf{x}_j + \mathbf{t}_{ee}(k)\,])$ \\
    $\mathbf{r}(\gamma,k) \gets [\,(\hat{\mathbf{u}}_j-\mathbf{u}_j)_{j=1}^3,\;\lambda_k k\,]^\top$ \\

    \CommentSty{// Robust optimization} \\
    $(\gamma^\star,k^\star) \gets \underset{\gamma,k}{\arg\min}\ \mathcal{L}(\mathbf{r}(\gamma,k))$ \\
    \quad \text{TRF solver with Cauchy loss $\mathcal{L}$} \\

    \CommentSty{// Compose final pose} \\
    $\mathbf{T}_{\mathrm{cam}\rightarrow\mathrm{ee}} \gets
        \begin{bmatrix}
        \mathbf{R}_{ee}(\gamma^\star) & \mathbf{t}_{ee}(k^\star) \\
        \mathbf{0}^\top & 1
        \end{bmatrix}$ \\

    \Return $\mathbf{T}_{\mathrm{cam}\rightarrow\mathrm{ee}}$
\end{algorithm}

As shown in Fig. \ref{training data}, the keypoint on the end-effector, Outer Roll,  $\mathbf{u}_0=(u_0,v_0)^\top$ corresponds to a 3D point at the end of the shaft centerline geometrically (i.e. it is on the centerline of the insertion shaft).
Its position in space can be obtained by calculating the intersection point of the camera ray and the shaft centerline:

\begin{equation}
    \mathbf{r} = \frac{\mathbf{K}^{-1}[u_0,v_0,1]^\top}{\|\mathbf{K}^{-1}[u_0,v_0,1]^\top\|}
\end{equation}

\begin{equation}
    (\lambda^\star,\mu^\star) 
    = \arg\min_{\lambda,\mu} 
      \;\big\|\,\lambda\,\mathbf{r} - (\mathbf{a} + \mu\,\mathbf{d})\,\big\|^2,
\end{equation}
where $\mathbf{K}\in\mathbb{R}^{3\times 3}$ is the camera intrinsic,  
$\mathbf{r}\in\mathbb{R}^3$ is the unit ray direction passing through $(u_0,v_0)$,  and $\lambda,\mu\in\mathbb{R}$ are the ray and line parameters, respectively. The 3D position of this point can be recovered as
\begin{equation}
    \mathbf{p_0} = \mathbf{a} + \mu^\star\,\mathbf{d}
\end{equation}
hence providing the 3D position of the end of the insertion shaft.

The direction of the recovered centerline, $\textbf{d}$, also provides the information on the pitch and yaw of the end-effector (i.e. two rotational degrees of freedom).
We compute this as an alignment transform which is solved by Rodrigues’ formula: 
\begin{equation}
    \mathbf{R}_{\text{align}}
    = \mathbf{I}_3 + [\mathbf{v}]_\times 
      + [\mathbf{v}]_\times^2 \,\frac{1-c}{s^2}, 
      \qquad \text{for } s > 0.
\end{equation}
where 
\begin{equation}
    \small
    \mathbf{z} = [0,0,1]^\top, 
    \qquad 
    \mathbf{v} = \mathbf{z} \times \hat{\mathbf{d}}, 
    \qquad 
    s = \|\mathbf{v}\|, 
    \qquad 
    c = \mathbf{z}^\top \hat{\mathbf{d}}.
\end{equation}
and skew–symmetric matrix of $\mathbf{v}$ is defined as 
\begin{equation}
    [\mathbf{v}]_\times =
    \begin{bmatrix}
    0 & -v_3 & v_2 \\
    v_3 & 0 & -v_1 \\
    -v_2 & v_1 & 0
    \end{bmatrix}.
\end{equation}

\begin{figure*}[ht]
    \centerline{\includegraphics[width=0.90\linewidth,clip=true,trim={2mm 2mm 2mm 2mm}]{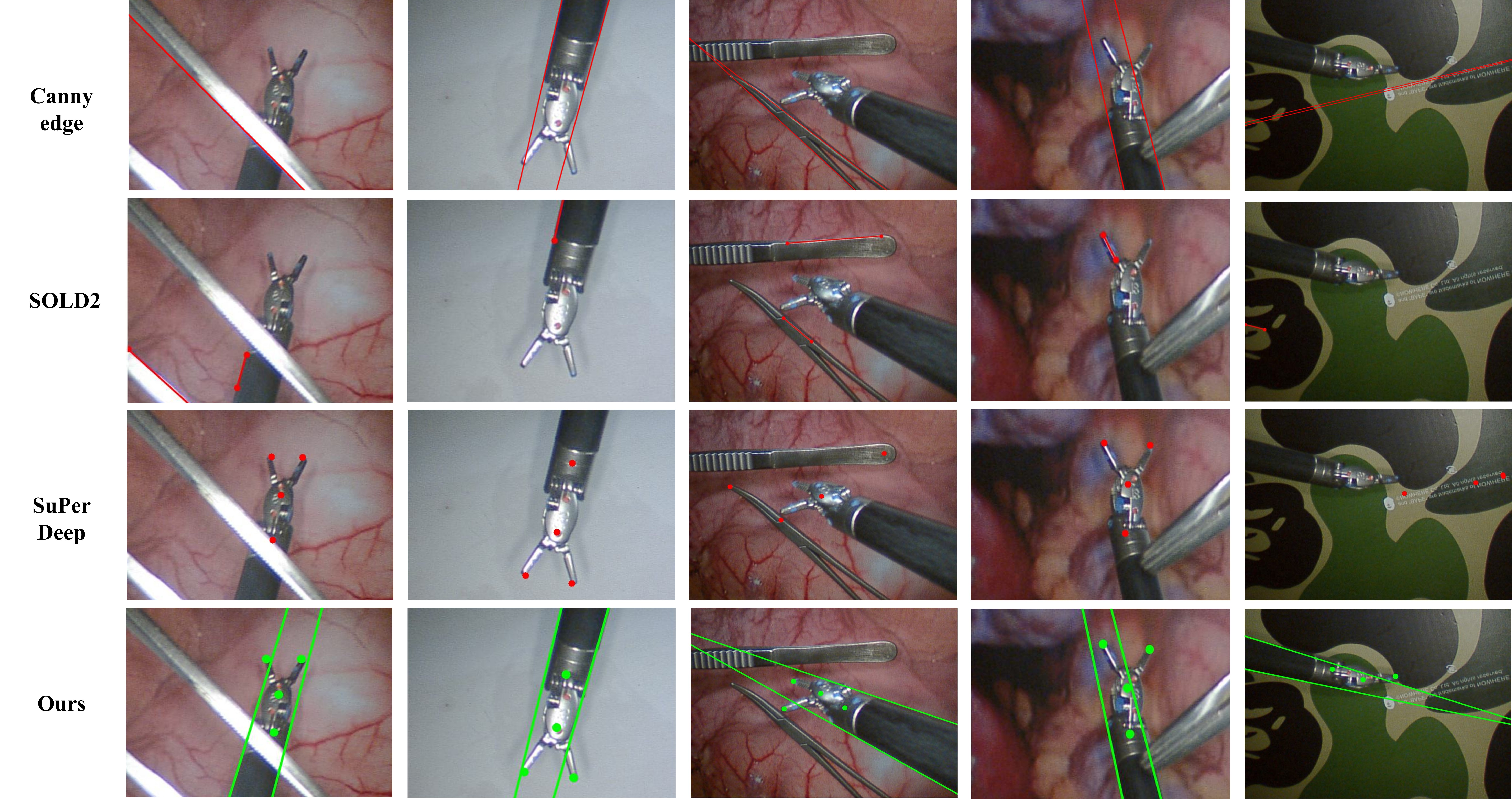}}
    \caption{Qualitative comparison of feature detection results between our and prior models. Prior models follow the same implementation as in the original papers.
    }
    \label{qualitative}
    \vspace{-0.14in}
    
\end{figure*}

\subsubsection{Solving for shaft roll}
The previously recovered information, the end-effector's pitch and yaw, $\mathbf{R}_{\text{align}} $ and position $\mathbf{p_0}$, are used as an initial guess of the end-effector pose. As discussed in prior sections, keypoint features are susceptible to noise and unreliable in real-world conditions. To address this challenging issue, the final pose is constructed by solving for two more factors:
\begin{equation}
    \mathbf{R}_{ee} = \mathbf{R}_{\text{align}}\mathbf{R}_z(\gamma),
    \qquad 
    \mathbf{t}_{ee} = \mathbf{p}_0 + k\,{\mathbf{d}},
\end{equation}
where $\gamma$ is the rotation angle around the shaft orientation (i.e. the missing rotational component about the end-effector) and $k$ is the scaling factor for compensating the noisy Outer Roll keypoint detection. With joint angle reading $\mathbf{q}$, the last three keypoints' positions $\mathbf{x}_j \in\mathbb{R}^3$ in the end-effector frame can be obtained using forward kinematics:
\begin{equation}
    \{\mathbf{x}_j\}_{j=1}^{3} = \text{FK}(\mathbf{q})
\end{equation}
providing 2D projections with $\gamma$ and $k$ as
\begin{equation}
    \widehat{\mathbf{u}}_j(\gamma,k) = 
    \pi\!\big(\mathbf{K}[\mathbf{R}_{ee}\mathbf{x}_j+\mathbf{t}_{ee}]\big).
\end{equation}
We construct a reprojection residual vector,
\begin{equation}
    \mathbf{r}(\gamma,k) =
    \left[
        \big(\widehat{\mathbf{u}}_j(\gamma,k)-\mathbf{u}_j\big)_{j=1}^J,\;\lambda_k k
    \right]^\top,
\end{equation}
where $\lambda_k$ denotes the regularization weight penalizing keypoint drift in the optimization, to provide a loss,
\begin{equation}
    (\gamma^\star,k^\star) = \arg\min_{\gamma,k}\,\mathcal{L}(\mathbf{r}(\gamma,k)),
\end{equation}
which will be optimized using a Trust-Region Reflective (TRF) solver with robust Cauchy loss to cope with feature outliers.
The final end-effector pose can be constructed as
\begin{equation}
    \mathbf{T}_{\mathrm{cam}\rightarrow\mathrm{ee}} =
    \begin{bmatrix}
    \mathbf{R}_{ee}(\gamma^\star) & \mathbf{t}_{ee}(k^\star)\\
    \mathbf{0}^\top & 1
    \end{bmatrix}.
\end{equation}
While achieving robust performance against noise, this pose solver contributes negligibly to the overall runtime of the pipeline due to its simplicity and low dimensionality.


\section{Experiments}

\subsection{Implementation setups}

The framework is implemented in PyTorch and trained on an NVIDIA RTX 3090 GPU. The synthetic training set consists of 20{,}000 rendered frames with complete feature annotations. Training is performed with mixed precision using the AdamW optimizer (learning rate $2{\times}10^{-4}$, weight decay $10^{-4}$) and a cosine learning rate schedule with $1\%$ warm-up (\textit{get\_cosine\_schedule\_with\_warmup}). For inference, the pose parameters $(\gamma,k)$ are estimated via a least-squares solver from \textit{SciPy}, employing the Trust-Region Reflective method with a Cauchy loss. The parameter bounds are set to $\theta \in [-\pi,\pi]$ and $k \in [-0.015,0.015]$.

\begin{table*}[ht]
\small
\centering
\setlength{\tabcolsep}{6pt}
\begin{tabular}{c c cccccc c}
\toprule
\multirow{2}{*}{Method} & \multirow{2}{*}{Kpt / Edge} 
    & \multicolumn{2}{c}{Structured} 
    & \multicolumn{2}{c}{Distracted} 
    & \multicolumn{2}{c}{Occluded} 
    & \multirow{2}{*}{Time (ms)} \\
\cmidrule(lr){3-4}\cmidrule(lr){5-6}\cmidrule(lr){7-8}
 & & Kpt $\downarrow$  & Edge $\uparrow$ & Kpt $\downarrow$  & Edge $\uparrow$ &  Kpt $\downarrow$  & Edge $\uparrow$ &  \\
\midrule
Canny edge \cite{richter2021robotic}    & \ding{55} / \ding{51}& -- & 0.7995 & -- & 0.6774 & -- & 0.6166 & \textbf{2.87} \\
SOLD2 \cite{d2024robust} &  \ding{55} / \ding{51} & -- & 0.4291 & -- & 0.3656 & -- & 0.3532 & 67.58 \\
SuPer Deep \cite{lu2021super}     & \ding{51} / \ding{55}  & 47.09 &  -- & 71.72 &  -- & \underline{31.12} &  -- & 29.37 \\

Ours (only keypoints)      & \ding{51} / \ding{55}  & \textbf{15.37} & -- & \textbf{22.95} & -- & \textbf{23.39} & -- & 24.17 \\

Ours (only edges)      & \ding{55} / \ding{51} & -- & 0.9164 & -- & \textbf{ 0.9667} & -- & \textbf{0.9679} & 55.82 \\

Ours (w/o edge refinement)      & \ding{51} / \ding{51}  & 22.16 & 0.9168 & 34.20 & 0.9107 & 34.74 & 0.9216 & 30.31 \\

Ours (final)      & \ding{51} / \ding{51}  & \underline{22.16} & \textbf{0.9315} & \underline{34.20} & \underline{0.9291} & 34.74 & \underline{0.9478} & 61.79 \\

\bottomrule
\end{tabular}
\caption{Quantitative comparison results. Keypoint and edge accuracies are evaluated with per-keypoint localization error (in pixels) and modified EA score, respectively. “Only keypoints/edges” denote ablation variants with solely the Keypoint Net or Edge Net trained on the backbone. The best results are shown in \textbf{bold}, and the second best are \underline{underlined}. Reported times indicate the average per-frame inference latency (ms) for feature detection.}
\label{quantitative}
\vspace{-0.1in}
\end{table*}

\subsection{Feature detection under real-world conditions}
Surgical robot feature detection is a crucial yet challenging task as a result of the dynamic and uncertain nature of real-world environments, which has significantly limited the performance of previous approaches. In this section, comprehensive evaluation and analysis are presented, summarizing prior approaches and evaluating them against the unified feature detection network on noisy real-world data. The evaluation dataset contains 290 frames of real images in diverse scenes with manual labeling and refinement.  It is divided into three categories based on environmental conditions: \textit{Structured} (100 frames), which contain only the surgical robot arms against randomized backgrounds; \textit{Distracted} (114 frames), which introduce additional surgical instruments or visual clutter in the scene; and \textit{Occluded} (76 frames), where parts of the instrument are partially hidden by tools or other obstructions.

\textbf{Qualitative results}
The qualitative comparison is demonstrated in Fig.\ref{qualitative}), where detected features for each model are overlaid on the raw images. The result comprises feature detection output of the proposed model and prior approaches, including Canny edge detection deployed in \cite{richter2021robotic}, SOLD2 \cite{d2024robust}, and DeepLabCut \cite{mathis2018deeplabcut} from SuPer Deep \cite{lu2021super}. All the previous models are implemented following the original setup. With jointly learned features, the proposed model can accurately output keypoints and shaft edges in both clean and clustered environments. In contrast, the baseline methods usually struggle due to their reliance on either low-level edge operators (Canny), label matching across diverse frames (SOLD2), or keypoint-only network output (DeepLabCut), leading to incomplete or unstable detections under complex surgical conditions.

\textbf{Quantitative results} The qualitative performance comparison across three scene categories is presented in Table~\ref{quantitative}. In addition to the baselines adopted in previous works, ablation variants of the proposed method are reported to further evaluate and analyze the contribution of each component to the overall performance. All trainable models are trained on the same synthetic dataset for benchmarking. 

Quantitatively, each model is evaluated on feature detection accuracy and network runtime. Keypoint performance is measured using the per-keypoint localization error, defined as the Euclidean distance between predicted and ground-truth keypoints:
\vspace{-0.08in}
\begin{equation}
    \text{Err}_{\text{kpt}} = 
    \frac{1}{N J} \sum_{i=1}^{N} \sum_{j=1}^{J} 
    \big\| \hat{\mathbf{u}}_{ij} - \mathbf{u}_{ij} \big\|_2 ,
\end{equation}
where $N$ is the total frames number and $J$ is the keypoints number.

Edge detection accuracy is evaluated using a modified EA-score, following \cite{zhao2021deep}. Given two predicted lines $\{\textbf{e}_i\}_{i=1}^2$ and reference lines $\{\textbf{e}_i^{\ast}\}_{i=1}^2$ with association ambiguity, 
the Edge Agreement (EA) score can be calculated as
\begin{equation}
    \mathrm{EA} = \mathrm{SE} \cdot \mathrm{SA},
\end{equation}
where the spatial extent (SE) term measures the agreement of line segment centers
\begin{equation}
    \mathrm{SE} = 1 - 
    \frac{\sqrt{d_{\mathrm{Chamfer}}(\{\textbf{c}_i\},\{\textbf{c}_i^{\ast}\})}}
         {\sqrt{H^2+W^2}},
\end{equation}
with $d_{\mathrm{Chamfer}}$ denoting the Chamfer distance between point sets, $\textbf{c}_i$ and $\textbf{c}_i^{\ast}$ the segment midpoints of predicted and reference lines, 
and $H,W$ the image height and width. The structural agreement (SA) term measures angular consistency:
\vspace{-0.1in}
\begin{equation}
    \mathrm{SA} = 1 - 
    \frac{\min\!\big(\Delta\theta_{(1\!\to\!1,2\!\to\!2)},
                     \Delta\theta_{(1\!\to\!2,2\!\to\!1)}\big)}
         {\tfrac{\pi}{2}},
\end{equation}
where $\Delta\theta_{(\cdot)}$ is the mean absolute angle difference between paired lines. 
Thus, EA attains $1$ for perfectly aligned edges and decreases with increasing spatial or angular deviation.

In Table.\ref{quantitative}, feature detection accuracy and inference run time for each model are reported. The quantitative comparison highlights the effectiveness of the introduced method across diverse scenes, which outperforms the previous approaches by a great margin in both keypoint and edge detections. The keypoint-only and edge-only variants achieve the best performance within their respective categories by fully utilizing the backbone network, while the combined network enables joint feature detection without introducing excessive performance loss or significant additional runtime.

\begin{table}[t]
    \footnotesize
    \centering
    \begin{tabular}{l p{1.4cm} p{1.4cm} p{1.4cm} p{1.2cm}}
        \toprule
        Method & Easy  & Medium  & Hard & time \\
        \midrule
        PnP & 0.05456 & 0.06173 & 0.06761 & 6.51 (s)\\
        \cite{liang2025differentiable}  & 0.00046 & 0.00243 & 0.00253 & 670.67 (s)\\
        Ours & \textbf{0.00032} & \textbf{0.00132} & \textbf{0.00095} & 0.072 (s)\\
        \bottomrule    
    \end{tabular}
    \caption{RCM convergence comparion of the proposed model and previous approaches. The results are in meters.}
    \label{consistency}
    \vspace{-0.17in}
\end{table}

\subsection{Pose reconstruction accuracy}

Surgical robot arms typically operate around a fixed Remote Center of Motion (RCM) as a physical constraint.  
Liang et al.~\cite{liang2025differentiable} introduced an efficient method to evaluate the quality of surgical robot pose reconstruction by calculating the spatial convergence of the calibrated poses. As the PSM rotates its insertion shaft around a fixed RCM point, an ideal calibration should yield cylinder axes that intersect at a unique converging point in 3D space. This point is estimated by minimizing the sum of squared distances to all recovered cylinder axes:
\vspace{-0.1in}
\begin{equation}
    \mathbf{x}^* = \arg \min_{\mathbf{x} \in \mathbb{R}^3} 
    \sum_{i=1}^{N} \left\| (\mathbf{x} - \mathbf{p}_i) - (\mathbf{d}_i^\top(\mathbf{x} - \mathbf{p}_i)) \mathbf{d}_i \right\|^2,
\end{equation}
where $\mathbf{p}_i$ and $\mathbf{d}_i$ denote the origin and direction of the $i$-th shaft. The standard deviation of distances from $\mathbf{x}^*$ to each axis is used to measure the calibration consistency. Meanwhile, the process time of each method is reported to quantitatively evaluate the pose reconstruction efficiency. 

Table~\ref{consistency} reports results across the dataset of three calibration difficulty levels. Compared to the PnP solver \cite{richter2021robotic} and the differentiable rendering approach \cite{liang2025differentiable}, the proposed framework achieves substantially lower standard deviation in all cases, demonstrating robustness and consistency across diverse scenarios. Moreover, the PnP approach relies on manual annotation and point association, while the differentiable rendering method involves iterative optimization that can take hundreds of seconds per frame. In contrast, our approach only takes milliseconds to complete a full forward pass.

Additionally, the pose estimation results of the proposed framework and the differentiable rendering approach are visualized in Fig.~\ref{vs diff}, where the projected tool skeleton is overlaid on the original images. While the differentiable rendering-based method depends heavily on the quality of silhouette masks and the stability of optimization for robustness, our framework directly bridges the extracted features to the robot pose without costly iterative refinement, achieving both higher accuracy and substantially faster inference.


\section{Discussions and conclusion}

In this work, we present a robust pose estimation framework for surgical robot instruments using a unified feature detection network. By unifying shaft edges and keypoints as jointly learnable features, the method delivers reliable detection across diverse environmental conditions. The proposed framework incorporates an efficient geometry-based pose inference pipeline that directly bridges the feature-to-pose gap, effectively overcoming the long runtimes and convergence issues of prior approaches. In the future, we plan to extend the framework to dual-arm configurations and multiple instrument categories, and to further address occlusion challenges through the integration of filter-based techniques.





\balance
\bibliographystyle{ieeetr}
\bibliography{references}

\end{document}